\documentclass[review]{elsarticle}
\graphicspath{ {./figures/} }
\usepackage{hyperref}
\usepackage{float}
\usepackage{verbatim} %comments
\usepackage{apalike}
\usepackage{amsfonts}
\usepackage{amsmath}
\usepackage{adjustbox}
\usepackage{graphicx}
\usepackage{makecell}
\usepackage{booktabs}
\usepackage{placeins}
\usepackage{dblfloatfix}
\restylefloat{figure}
\floatstyle{plaintop} %table caption at top
\restylefloat{table}
\usepackage{caption}
\journal{ }

\biboptions{authoryear}

\begin{document}
\begin{frontmatter}

%The title page must contain the title of the paper and the full name/full affiliation with country/e-mail address for each author and co-author of the manuscript. Please make sure you have included all elements listed below with your manuscript submission.

\begin{titlepage}
\begin{center}
\vspace*{1cm}

\textbf{ \large Improving Adversarial Robustness of Zero-Shot CLIP with Confidence-Aware Weighting}

\vspace{1.5cm}

% Author names and affiliations
Nikoo Naghavian$^a$ (nikoo.naghavian@ut.ac.ir), Mostafa Tavassolipour$^a$ (tavassolipour@ut.ac.ir)\\

\hspace{10pt}

\begin{flushleft}
\small  
$^a$ School of ECE, College of Engineering, University of Tehran, Tehran, Iran 

\begin{comment}
Clearly indicate who will handle correspondence at all stages of refereeing and publication, also post-publication. Ensure that phone numbers (with country and area code) are provided in addition to the e-mail address and the complete postal address. Contact details must be kept up to date by the corresponding author.
\end{comment}

\vspace{1cm}
\textbf{Corresponding author: } \\
Mostafa Tavassolipour \\
School of ECE, College of Engineering, University of Tehran, Tehran, Iran \\
Email: tavassolipour@ut.ac.ir

\end{flushleft}        
\end{center}
\end{titlepage}

\title{Improving Adversarial Robustness of Zero-Shot CLIP with Confidence-Aware Weighting}

\author[label1]{ Nikoo Naghavian}
\ead{nikoo.naghavian@ut.ac.ir}

\author[label1]{Mostafa Tavassolipour\corref{cor1}}
\ead{tavassolipour@ut.ac.ir}

\cortext[cor1]{Corresponding author.}
\address[label1]{School of ECE, College of Engineering,\\
  University of Tehran, Tehran, Iran}
 \address[label2]{School of ECE, College of Engineering\\
   University of Tehran, Iran}

\begin{abstract}
Vision-language models such as CLIP demonstrate impressive zero-shot generalization but remain highly vulnerable to adversarial attacks. Prior adversarial methods treat all samples equally in the loss function, despite the fact that not all inputs contribute equally to adversarial vulnerability. Some samples retain high confidence even under large perturbations, while others are unstable and change predictions with minor noise. To address this, we propose Confidence-Aware Weighting (CAW) to enhance the zero-shot robustness of vision-language models. CAW introduces two key components: (1) a Confidence-Aware Loss that prioritizes uncertain adversarial examples by scaling the KL divergence between clean and adversarial predictions, and (2) a Feature Alignment Regularization that preserves semantic consistency by minimizing the distance between frozen and fine-tuned image encoder features on adversarial inputs. Together, these components improve both clean and robust accuracy without compromising generalization. Extensive experiments on TinyImageNet and 14 additional datasets show that CAW outperforms recent state-of-the-art methods such as PMG-AFT and TGA-ZSR under strong adversarial attacks like AutoAttack, while also being more memory-efficient.
\end{abstract}


\end{frontmatter}

\section{Introduction}
\label{introduction}
Traditional deep learning approaches rely on pre-training followed by fine-tuning with labeled data for each downstream task. The emergence of GPT-3 \citep{mann2020language} in the natural language processing field has popularized models with zero-shot capability, where models trained on diverse internet-scale data can be applied to a wide range of tasks and unseen domains. In the multimodal setting, CLIP \citep{radford2021learning} employs a contrastive loss \citep{chen2020simple, caron2021emerging} to align matching image–text pairs in a shared embedding space while separating mismatched pairs. This enables the model to acquire broad vision–language knowledge and achieve strong performance across various tasks, including image classification \citep{simonyan2014very, he2016deep}, semantic segmentation \citep{chen2017deeplab}, object detection \citep{girshick2015fast, he2020momentum}, image–text retrieval \citep{cao2022image}, and visual question answering \citep{parelli2023clip}. Although CLIP demonstrates strong generalization ability, it remains vulnerable to small, imperceptible perturbations that leave the image visually unchanged to humans but cause significant shifts in predictions \citep{goodfellow2014explaining}. 
% Therefore, enhancing the robustness of CLIP is essential for safety in real-world and high-risk applications~\citep{jia2024bev, liu2023clip}.
Adversarial training \citep{shafahi2019adversarial, WANG2022162} is among the most effective approaches for improving robustness against strong attacks,  
% such as PGD \citep{madry2017towards}, AutoAttack \citep{croce2020reliable}, and CW \citep{carlini2017towards} 
typically training from scratch with both adversarial and clean examples. However, when applied to large-scale models like CLIP, adversarial training must be adapted to prevent overfitting and the forgetting of pre-trained knowledge, while still enhancing robustness \citep{zhao2023evaluating, zhang2019theoretically, mao2022understanding}.

The TeCoA \citep{mao2022understanding} method was the first to study the zero-shot robustness of large-scale vision-language models. It showed the importance of using text supervision with a contrastive adversarial loss while applying different adaptation approaches \citep{bahng2022exploring}. Later, the PMG-AFT \citep{wang2024pre} method added new terms to the previous loss function to enhance robustness while causing a smaller decrease in performance on clean data. More recently, TGA-ZSR \citep{yu2024text} introduced a method that improves both robustness and clean accuracy, along with the interpretability of attacks. This approach used text supervision with semantic information instead of relying on the model’s output probabilities. Despite their effectiveness, these methods either need high memory usage, or still struggle to maintain robust accuracy under strong attacks.

However, existing methods treat all adversarial examples equally during training, regardless of their difficulty or stability. This assumption limits robustness since samples with different confidence levels contribute unequally to model learning. Here, we define confidence as the model’s predicted probability for the true class. As observed in our analysis, samples with low robust confidence are more likely to change their predicted labels under perturbations, indicating that the model is less stable on these examples. On the other hand, highly confident clean samples are already robust and require less adaptation. Therefore, applying the same weight to all examples can lead to inefficient optimization and overfitting on easy samples. To address this issue, our method introduces a confidence-aware weighting mechanism that assigns larger weights to low-confidence adversarial examples, allowing the model to focus more on unstable samples and improve robustness without sacrificing clean accuracy.

We propose a adversarial fine-tuning loss named Confidence-Aware Weighting (CAW) that improves the robustness of a pre-trained CLIP model while preserving clean accuracy and reducing memory usage. To the best of our knowledge, this combination has not been previously explored for improving adversarial robustness in zero-shot VLMs. This method introduces two key components designed to improve robustness and maintain generalization. The first is a Confidence-Aware term, which weights the KL divergence between clean and adversarial prediction distributions of the fine-tuned and frozen pre-trained CLIP models, ensuring that training focuses more on hard adversarial examples. The second is a regularization term, which matches adversarial image features from the fine-tuned image encoder with those from the frozen pre-trained encoder, helping retain semantic knowledge from the pre-trained model and reducing overfitting. Experiments on TinyImageNet and 14 zero-shot datasets (see Section~\ref{subsec:datasets} for details) demonstrate state-of-the-art performance under AutoAttack, surpassing both PMG-AFT and TGA-ZSR in robust accuracy. Under PGD-100 and CW, the proposed method outperforms PMG-AFT in both robust and clean accuracy, while maintaining lower memory usage than both baselines.

The key contributions of this work are:
\begin{itemize}
    \item We propose a Confidence-Aware Weighting method that focuses on challenging samples by scaling the KL divergence term to improve zero-shot robustness while maintaining better retention of prior knowledge on clean inputs.
    \item It demonstrates superior robust performance compared to both PMG-AFT and TGA-ZSR under AutoAttack.
    \item It also shows improvements in both clean and robust accuracy over PMG-AFT under PGD-100 and CW attacks.
    \item Furthermore, it uses less memory compared to PMG-AFT and TGA-ZSR.
\end{itemize}
\section{Related Work}
\label{subsec:related}
\paragraph{Adversarial robustness}  
Deep neural networks have achieved remarkable performance on complex tasks, often producing highly confident predictions. However, small, imperceptible perturbations to the input can easily mislead them, resulting in incorrect outputs~\citep{madry2017towards, moosavi2016deepfool, szegedy2013intriguing, goodfellow2014explaining, kurakin2018adversarial}. To address this vulnerability, various techniques have been proposed, including distillation~\citep{linardatos2020explainable}, model compression~\citep{liu2018security}, activation pruning~\citep{dhillon2018stochastic}, gradient regularization~\citep{gu2014towards, ross2018improving} and adversarial training~\citep{madry2017towards}. Adversarial training remains the most effective approach, augmenting adversarial examples alongside clean data during training to improve robustness while maintaining generalization~\citep{ilyas2019adversarial}. Methods such as TRADES~\citep{zhang2019theoretically} balance clean accuracy and robustness by combining standard classification loss with a robustness regularization term. MART~\citep{wang2019improving} highlights the importance of misclassified examples for improving robustness, while ARoW~\citep{yang2023improving} focuses on the most vulnerable samples to enhance both generalization and robustness. HAT~\citep{rade2022reducing} mitigates over-robustness by introducing helper examples, arguing that pushing decision boundaries too far can harm clean accuracy.
\paragraph{Zero-shot Adversarial Robustness for VLMs}  
The introduction of the attention mechanism~\citep{vaswani2017attention}, combined with advances in GPUs and access to large-scale unlabeled internet data, enabled the development of language models like BERT~\citep{devlin2019bert}, GPT-2~\citep{radford2019language}, and GPT-3~\citep{mann2020language}, marking a new era in deep learning. GPT-3’s emergence brought zero-shot capabilities, allowing knowledge transfer to unseen domains and tasks. Following this trend, vision-language models (VLMs) such as BLIP~\citep{li2022blip}, CLIP~\citep{radford2021learning} and ALIGN~\citep{jia2021scaling} incorporate textual information with images to improve performance across diverse tasks rather than a single downstream application. Despite their generalization ability, VLMs remain vulnerable to imperceptible perturbations in the input, which can cause incorrect predictions\citep{zhao2023evaluating}. Recent research has explored enhancing VLM robustness. TeCoA~\citep{mao2022understanding} introduced the use of text knowledge for model alignment with adversarial examples through contrastive loss. PMG-AFT~\citep{wang2024pre} and TGA-ZSR~\citep{yu2024text} extended this approach by adding terms such as KL divergence or semantic alignment with text embeddings to improve both clean and robust accuracy. Another method~\citep{li2024language} extracts normalized semantic feature embeddings (anchors) for each class label from a CLIP text encoder and uses them to guide the image encoder during adversarial training, enabling robustness transfer to unseen categories. FARE approach~\citep{schlarmann2024robust} aligns adversarial example features directly with the embeddings of a pre-trained CLIP model without requiring labels. Another related work~\citep{dongimproving} leverages not only the final adversarial examples from the PGD process but also intermediate samples along the adversarial trajectory for training. Unlike FARE, which aligns clean frozen representations with adversarial fine-tuned representations, our method aligns adversarial representations from both frozen and fine-tuned encoders. Furthermore, it combines feature-level alignment with confidence-aware weighting at the prediction level, enabling the study of their complementary effects on zero-shot robustness.

\section{Methodology}
% \begin{figure}[h]
%     \centering
%     \includegraphics[width=1\textwidth]{pipline.pdf}
%     \caption{Overview of Confidence-Aware Weighting (CAW) method. $\odot$ means matrix inner product. 
%     }
%     \label{fig:schema}
% \end{figure}
\subsection{Preliminaries and Problem Setup}
In this work, we employ CLIP \citep{radford2021learning} to enhance zero-shot robustness in classification tasks. CLIP has two encoders that learn a joint visual–text feature space. At inference, the predicted label is the one that
text embedding has the highest cosine similarity with the image embedding. As in the original CLIP setup, we use text embeddings obtained by encoding class names through a fixed prompt template: \texttt{"This is a photo of a [class]"}. Following prior works \citep{yu2024text, wang2024pre}, we fine-tune the image encoder using the cross-entropy loss:
\begin{equation}
L_{CE}(x, t, y) = -\mathbb{E}_{i,j} \left[ y_{ij} \log \frac{\exp(\cos(f(x)_i, g(t)_j)/\tau)}{\sum_k \exp(\cos(f(x)_i, g(t)_k)/\tau)} \right]
\label{eq:ce_loss}
\end{equation}
where \( f(x) \) and \( g(t) \) denote the image and text embeddings, \(\tau\) is the temperature parameter, and \(\cos\) is the cosine similarity. The label \( y_{ij} \) is set to 1 for positive image-text pairs and 0 otherwise.
\subsubsection{Adversarial Attacks}  
Deep learning models are typically trained and evaluated on clean data; however, small, imperceptible perturbations can cause significant prediction errors. Such perturbations can be generated using attack methods including PGD \citep{madry2017towards}, AutoAttack \citep{croce2020reliable}, and CW \citep{carlini2017towards}. PGD is an iterative attack that applies noise over multiple steps, producing stronger adversarial examples \( x_a \) than single-step methods such as FGSM \citep{goodfellow2014explaining}. It seeks perturbations that maximize the loss while keeping the perturbed input within a specified neighborhood of the clean example:
\begin{equation}
x_{a+1} = \Pi_{x+\mathcal{S}}\left(x_a + \varepsilon \cdot \operatorname{sign}\left(\nabla_{x_a} L(x_a, t, y)\right)\right)
\label{eq:pgd_update}
\end{equation}
where \( L \) denotes the loss function, \( x \) is the clean input, \( \varepsilon \) is the perturbation bound under the \( p \)-norm, and \( \nabla_x L \) is the gradient direction that increases the loss. The set \( \mathcal{S} \) represents the allowed changes that the adversary can make to the input.

\subsubsection{Adversarial Training}
By optimizing over adversarially perturbed inputs, adversarial fine-tuning enables models to learn more robust features through a min-max objective. In some cases, including our method, the objective function used to craft adversarial examples differs from the one used to optimize the model parameters. Specifically, in the inner loop, adversarial examples \( x_a \) are generated by maximizing the loss \( L \) (i.e., \( \mathcal{L}_{\text{CE}} \), as defined in Equation~\ref{eq:ce_loss} of our method), which is optimized using the PGD update rule (Equation~\ref{eq:pgd_update}):
\begin{equation}
	x_a = \arg\max_{\|x_{a}-x\|_\infty \le \epsilon}
	L(x,t,y).
	\label{eq:inner_loop}
\end{equation}
In the outer loop, the model parameters \( \theta \) are updated by minimizing a separate loss function \( \mathcal{J} \) (i.e., \( \mathcal{L}_{\text{total}} \), as defined in Equation~\ref{eq:total} of our method):
\begin{equation}
	\theta^\star = \arg\min_{\theta} \; \mathcal{J}_{\theta}(x_a, t, y)
	\label{eq:outer_loop}
\end{equation}
where \( x_a \) denotes the adversarial example generated under the standard
\( \ell_\infty \)-bounded threat model, satisfying
\( \|x_a-x\|_\infty \le \epsilon \).

\subsubsection{Zero-Shot Adversarial Robustness}
Based on prior works, we use the Tiny-ImageNet dataset to fine-tune the CLIP model with an adversarial loss function, aiming to evaluate its performance on unseen adversarial samples during inference. Specifically, we measure the zero-shot accuracy of the fine-tuned model across various datasets using adversarial examples generated via white-box attacks.
\subsection{Method}
% put this before the paragraph that first mentions it
\begin{figure*}[t]
  \centering
  \includegraphics[width=\textwidth]{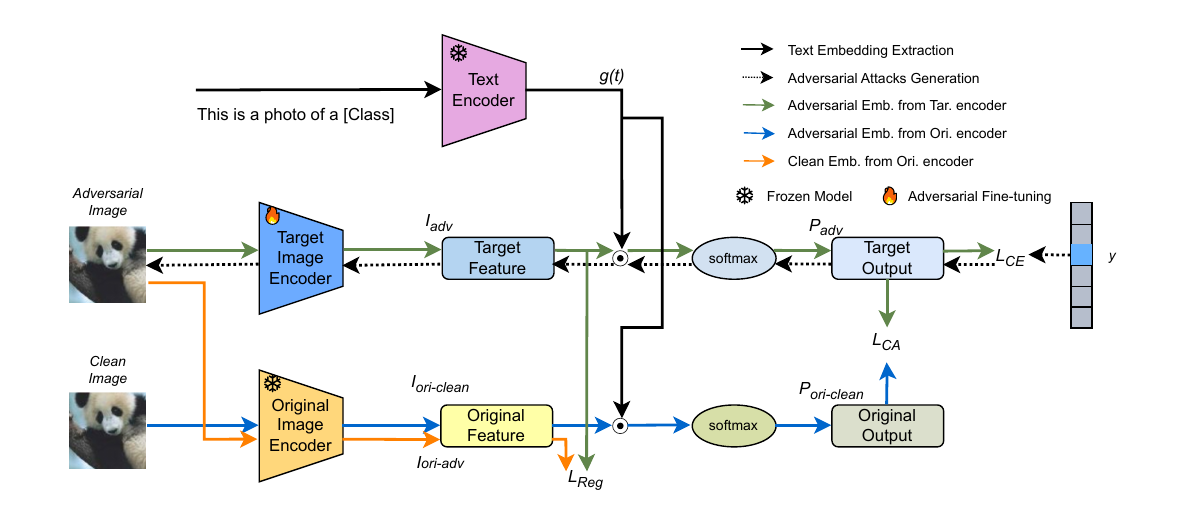}% no 1\textwidth
  \caption{Overview of the Confidence-Aware Weighting (CAW) method. $\odot$ denotes the matrix inner product. In this framework, only the target image encoder is fine-tuned. The method introduces two key loss terms: $\mathcal{L}_{CA}$, which encourages the distribution of adversarial features to align with the clean features from the pre-trained CLIP model, and $\mathcal{L}_{Reg}$, which preserves generalization by aligning adversarial features from the target and original (frozen) encoders.}
  \label{fig:schema}
\end{figure*}
\label{subsec:method}
Building on previous studies~\citep{mao2022understanding, wang2024pre, yu2024text}, we aim to preserve the generalizable and robust features learned by the pre-trained CLIP model during fine-tuning with a new loss function. As illustrated in Figure~\ref{fig:schema}, we use both the original and target image encoders to retain prior knowledge while improving robustness. Although the TeCoA method~\citep{mao2022understanding} introduces a contrastive loss using adversarial examples with text supervision, it remains insufficient for jointly improving clean and robust accuracy. To address this limitation, we propose two additional loss terms that enhance robustness while maintaining generalization to unseen tasks. 
\paragraph{Confidence-Aware Term}
We propose Confidence-Aware loss that focuses on challenging samples by emphasizing hard adversarial examples, i.e., those where the model is less confident in the correct class, while down-weighting easier ones. In contrast to prior methods that treat all samples equally in the loss function, our approach explicitly targets the inherent weaknesses in adversarial training by assigning more weight to samples that are more easily fooled by adversaries. This idea is inspired by the ARoW method~\citep{yang2023improving}, which prioritizes vulnerable samples to enhance adversarial robustness. However, our formulation differs in both design and scope, as it is tailored to the unique challenges of vision-language models and zero-shot generalization. Rather than introducing a new instance-reweighting principle, our goal is to investigate how confidence-aware weighting can be adapted to preserve zero-shot transferability and adversarial robustness in VLMs. Specifically, we define a KL-based alignment between the frozen CLIP model’s predictions on clean images, \( P^{\text{clean}} \), and the fine-tuned model’s predictions on adversarial images, \( P^{\text{adv}} \). This alignment allows the model to retain semantic knowledge from pre-training while learning to handle difficult adversarial examples. Unlike ARoW, which uses the reverse KL divergence ($\text{KL}(P^\text{clean} \| P^\text{adv})$), we  place the adversarial distribution as the first argument, i.e., $\text{KL}(P^{\text{adv}} \| P^{\text{clean}})$. The distributions \( P^{\text{adv}} \) and \( P^{\text{clean}} \) are defined as:
\begin{align}
P^{\text{adv}} &= \text{softmax}(f(x_{\text{adv}})_{\text{tar}} \cdot g(t)^\top), \\
P^{\text{clean}} &= \text{softmax}(f(x_{\text{clean}})_{\text{ori}} \cdot g(t)^\top),
\end{align}
where \( f(\cdot) \) and \( g(\cdot) \) denote the image and text encoder embeddings, and the dot operator represents the matrix inner product between these embeddings. The subscripts \texttt{tar} and \texttt{ori} refer to features from the fine-tuned and frozen image encoders, respectively. The element \( P^{\text{adv}}_{i, y_i} \) denotes the predicted probability for the true label \( y_i \) under the adversarial input \( x^{\text{adv}}_i \), as defined in Equation~\ref{eq:weight_term}:
\begin{equation}
P^{\text{adv}}_{i, y_i} = \left[ \text{softmax}\left( f(x^{\text{adv}}_i) \cdot g(t)^\top \right) \right]_{y_i}.
\label{eq:weight_term}
\end{equation}
To incorporate this into training, we minimize the KL divergence between \( P^{\text{adv}} \) and \( P^{\text{clean}} \), scaled by \( 1 - P^{\text{adv}}_{i, y_i} \) to give greater importance to uncertain adversarial examples. 

This results in the Confidence-Aware loss:
\begin{equation}
L_{\text{CA}} = \frac{1}{N} \sum_{i=1}^{N} \left[ \text{KL}\left( P^{\text{adv}}_i \,\|\, P^{\text{clean}}_i \right) \left(1 - P^{\text{adv}}_{i, y_i} \right) \right].
\label{eq:confidence_loss}
\end{equation}
To quantify the divergence between the adversarial distribution $P_{\text{adv}}$ and the clean distribution $P_{\text{clean}}$, we employ Kullback-Leibler (KL) divergence in two configurations: $\text{KL}(P_{\text{clean}} \parallel P_{\text{adv}})$ and $\text{KL}(P_{\text{adv}} \parallel P_{\text{clean}})$. Our experiments and theoretical analysis demonstrate that $\text{KL}(P_{\text{adv}} \parallel P_{\text{clean}})$ more effectively enhances model robustness, despite the common use of $\text{KL}(P_{\text{clean}} \parallel P_{\text{adv}})$ in prior work. By using $\text{KL}(P_{\text{adv}} \parallel P_{\text{clean}})$, the model is encouraged to assign low probabilities to classes where $P_{\text{clean}}$ assigns low probabilities. Consequently, if $P_{\text{clean}}$ assigns a low probability to a class while $P_{\text{adv}}$ assigns a high probability, the resulting large loss penalizes the model, promoting robustness by discouraging predictions toward incorrect classes. In contrast, $\text{KL}(P_{\text{clean}} \parallel P_{\text{adv}})$ primarily aligns $P_{\text{adv}}$ with the high-probability regions of $P_{\text{clean}}$, which may limit robustness by overlooking low-probability regions. We further multiply the KL term by a confidence weight $(1 - P^{\text{adv}}_{i,y_i})$ 
to give more focus to difficult or uncertain samples. 
When the model is already confident, the weight is small; 
when it is uncertain, the weight is large, so the model focuses more on those unstable cases. 
\paragraph{Regularization Term}
We introduce a regularization loss that encourages consistency between the image encoder features of the frozen model, \( f(\cdot)_{\text{ori}} \), and the fine-tuned model, \( f(\cdot)_{\text{tar}} \), for adversarial inputs. This loss is computed before the text alignment stage, where the image features contain rich semantic information about the visual input. By aligning these features using the \( \ell_2 \) distance metric, the model retains the pre-trained CLIP knowledge and reduces the risk of overfitting during adversarial fine-tuning. The regularization loss is defined as:
\begin{equation}
L_{\text{Reg}} = \frac{1}{N} \sum_{i=0}^{N} \left\| f(x_{\text{adv}})_{\text{tar}} - f(x_{\text{adv}})_{\text{ori}} \right\|_2.
\end{equation}
Unlike FARE, which aligns clean frozen representations with adversarial fine-tuned representations, our regularization aligns adversarial representations from both frozen and fine-tuned encoders. This design stabilizes adversarial feature representations and mitigates representation drift during adversarial fine-tuning.
The overall loss function is formulated as follows:
\begin{equation}
\label{eq:total}
L_{\text{total}} = L_{\text{CE}} + \alpha \cdot L_{\text{CA}} + \beta \cdot L_{\text{Reg}}.
\end{equation}

\section{Datasets and Implementation}
\label{subsec:datasets}
\paragraph{Datasets}
To evaluate both clean and adversarial performance, we conduct extensive experiments on a diverse collection of image classification datasets. Our primary model, a pre-trained CLIP, is fine-tuned on the TinyImageNet \citep{deng2009imagenet} dataset. Evaluation is then performed not only on TinyImageNet but also on 14 additional datasets spanning five distinct domains. These include general object recognition benchmarks such as CIFAR-10 \citep{krizhevsky2009learning}, CIFAR-100 \citep{krizhevsky2009learning}, STL-10 \citep{coates2011analysis}, Caltech-101 \citep{fei2006one}, and Caltech-256 \citep{griffin2007caltech}; fine-grained classification datasets like OxfordPets \citep{6248092}, Flowers102 \citep{nilsback2008automated}, FGVCAircraft \citep{maji2013fine}, and StanfordCars \citep{krause20133d}; scene recognition via SUN397 \citep{xiao2010sun}; domain-specific datasets including Food101 \citep{bossard2014food}, EuroSAT \citep{helber2019eurosat}, and DTD \citep{cimpoi2014describing}; and one medical imaging dataset, PCAM \citep{Bejnordi2017}.
\paragraph{Implementation Details}
\label{subsec:impliment}
For implementation, we use the ViT-B/32 architecture as the backbone for the CLIP model and fine-tune it on adversarial examples generated from the TinyImageNet dataset. Adversarial examples for both training and evaluation are produced using PGD attacks under the $\ell_\infty$ norm. Training updates all image encoder parameters using SGD with a learning rate of $1 \times 10^{-4}$, momentum of 0.9, weight decay of 0, and a batch size of 128. All experiments use PGD with 2 iterations and a perturbation bound of $1/255$. For evaluation, we employ PGD-100, AutoAttack, and CW, each with a step size equal to the perturbation bound. We set the hyperparameters $\alpha = 2$ and $\beta = 1$ to balance clean and robust accuracy. To ensure fair comparison, we adopt settings consistent with prior studies, which also used an RTX 3090 GPU.
% =========================
% Table 1
% =========================
\begin{table*}[!htbp]
\centering
\caption{Zero-shot robust accuracy under AutoAttack with $\epsilon = 1/255$ on 15 datasets. Best results are in bold and second-best results are underlined.}
\label{tab:autoattack}

\scriptsize
\renewcommand{\arraystretch}{1.1}
\setlength{\tabcolsep}{3pt}

\begin{adjustbox}{max width=\textwidth}
\begin{tabular}{l*{16}{c}}
\toprule
\textbf{Methods} &
\rotatebox{40}{\makecell{Tiny-ImageNet}} &
\rotatebox{40}{\makecell{CIFAR-10}} &
\rotatebox{40}{\makecell{CIFAR-100}} &
\rotatebox{40}{\makecell{STL-10}} &
\rotatebox{40}{\makecell{SUN397}} &
\rotatebox{40}{\makecell{Food101}} &
\rotatebox{40}{\makecell{OxfordPets}} &
\rotatebox{40}{\makecell{Flowers102}} &
\rotatebox{40}{\makecell{DTD}} &
\rotatebox{40}{\makecell{EuroSAT}} &
\rotatebox{40}{\makecell{FGVC\\Aircraft}} &
\rotatebox{40}{\makecell{Caltech-101}} &
\rotatebox{40}{\makecell{Caltech-256}} &
\rotatebox{40}{\makecell{StanfordCars}} &
\rotatebox{40}{\makecell{PCAM}} &
\rotatebox{40}{\makecell{Average}} \\
\midrule
CLIP & 0.02 & 0.01 & 0.08 & 0.03 & 0.04 & 0.01 & 0.00 & 0.03 & 0.16 & 0.12 & 0.06 & 0.43 & 0.10 & 0.11 & 0.22 & 0.09 \\
FT-Clean & 0.08 & 0.03 & 0.01 & 0.91 & 0.09 & 0.04 & 0.06 & 0.03 & 0.48 & 0.02 & 0.03 & 1.38 & 0.66 & 0.03 & 0.03 & 0.26 \\
FT-Adv & \underline{50.48} & 37.55 & 20.39 & 69.14 & 16.25 & 11.23 & 33.91 & 18.54 & \textbf{19.95} & \underline{11.59} & 1.65 & 49.90 & 39.24 & 7.57 & \textbf{48.84} & 29.08 \\
TeCoA & 35.03 & 28.18 & 16.09 & 66.08 & 17.41 & 13.05 & 34.81 & 20.80 & 15.37 & 11.40 & 1.32 & 54.54 & 40.15 & 7.15 & 47.12 & 27.23 \\
FARE & 28.59 & 23.37 & 13.58 & 60.70 & 9.72 & 13.88 & 27.72 & 15.48 & 9.15 & 0.25 & 0.87 & 47.45 & 36.68 & 6.77 & 10.23 & 20.30 \\
PMG-AFT & 44.26 & \underline{44.12} & \underline{23.66} & \underline{73.90} & 19.63 & \underline{17.25} & 39.25 & 20.87 & 13.72 & \textbf{11.99} & 1.68 & \underline{60.57} & 44.25 & 9.59 & \underline{48.53} & 31.55 \\
TGA-ZSR & 49.45 & 40.53 & 22.38 & 72.06 & \textbf{20.36} & 15.58 & \underline{40.31} & \underline{21.43} & \underline{17.13} & 11.19 & \textbf{2.64} & 57.16 & \underline{45.68} & \underline{10.47} & 48.03 & \underline{31.63} \\
\midrule
CAW & \textbf{50.52} & \textbf{47.35} & \textbf{26.35} & \textbf{74.27} & \underline{19.64} & \textbf{20.50} & \textbf{41.89} & \textbf{21.61} & 16.80 & 11.11 & \underline{2.52} & \textbf{62.79} & \textbf{47.27} & \textbf{12.23} & 47.81 & \textbf{33.51} \\
\bottomrule
\end{tabular}
\end{adjustbox}
\end{table*}
\subsection{Main Results}
To evaluate our method, we compare against the reported results of CLIP, FT-Clean, FT-Adv, TeCoA, FARE~\citep{schlarmann2024robust}, PMG-AFT, and TGA-ZSR, as presented in the TGA-ZSR paper~\citep{yu2024text}. FT-Clean and FT-Adv are fine-tuned using clean and adversarial examples, both with contrastive loss.
\paragraph{AutoAttack}  
As shown in Table~\ref{tab:autoattack}, our method outperforms all compared approaches under AutoAttack. On average, it achieves a 2\% improvement in robust accuracy, demonstrating that the proposed training strategy learns transferable and more robust features resistant to this stronger attack. The model is trained with PGD-2 using a perturbation bound of $\epsilon = 1/255$ and evaluated on AutoAttack with the same perturbation bound. 
% See Appendix~\ref{subsec:related} for related work, Appendix~\ref{subsec:datasets} for implementation details and datasets, Appendix~\ref{subsec:ablation} for additional experiments and ablation studies, and Appendix~\ref{subsec:discussion} for limitations and broader impact.
\begin{figure}[htbp]
  \centering
  \includegraphics[width=0.8\columnwidth]{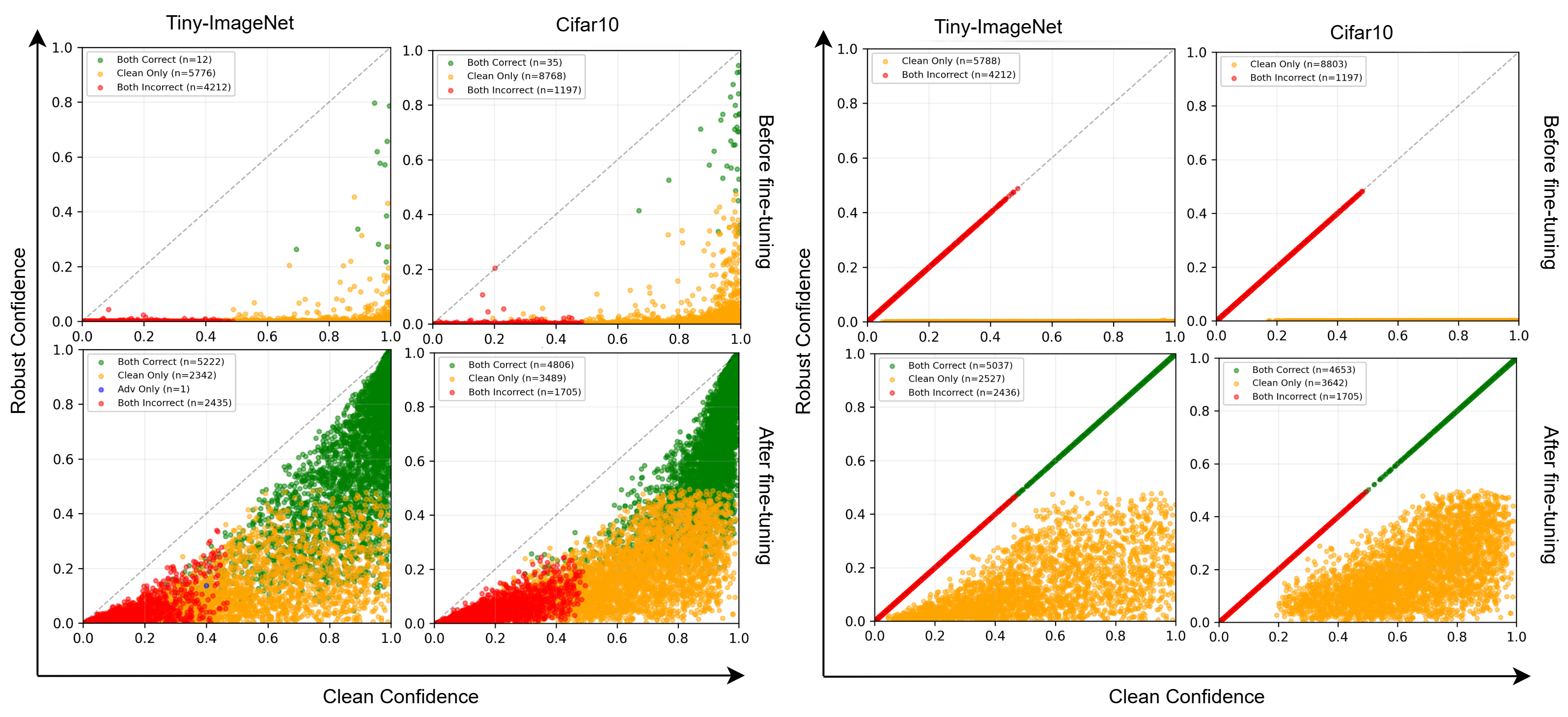}
  \caption{Scatter plots showing the relationship between clean confidence and robust confidence for samples from two datasets, before and after fine-tuning with CAW. The right plots correspond to PGD-100, and the left to AutoAttack.}
  \label{fig:confidence_pgd_aa}
\end{figure}
\label{subsec:ablation}
\paragraph{PGD and CW Attack}  
\label{subsec:PGD_CW}
As shown in Table~\ref{tab:robust_pgd100}, our method outperforms PMG-AFT in robust accuracy on most datasets, achieving a higher average performance. Table~\ref{tab:clean} further demonstrates that our method surpasses PMG-AFT in clean accuracy across all datasets. Based on these results, our method performs well on both clean and adversarial samples, showing competitive performance compared to other approaches. Table~\ref{tab:cw} indicates that our approach achieves better results than PMG-AFT under the CW attack. We compare only with CLIP and PMG-AFT because these are the only methods reported in the paper~\citep{yu2024text}. The model is trained with PGD-2 using a perturbation bound of $\epsilon = 1/255$ and evaluated on PGD-100 and CW with the same bound. 
% =========================
% Table 2
% =========================
\begin{table*}[!htbp]
\centering
\caption{Zero-shot robust accuracy under PGD-100 with $\epsilon = 1/255$ on 15 datasets, fine-tuned on TinyImageNet using PGD-2.}
\label{tab:robust_pgd100}

\scriptsize
\renewcommand{\arraystretch}{1.1}
\setlength{\tabcolsep}{3pt}

\begin{adjustbox}{max width=\textwidth}
\begin{tabular}{l*{16}{c}}
\toprule
\textbf{Methods} &
\rotatebox{40}{\makecell{Tiny-ImageNet}} &
\rotatebox{40}{\makecell{CIFAR-10}} &
\rotatebox{40}{\makecell{CIFAR-100}} &
\rotatebox{40}{\makecell{STL-10}} &
\rotatebox{40}{\makecell{SUN397}} &
\rotatebox{40}{\makecell{Food101}} &
\rotatebox{40}{\makecell{OxfordPets}} &
\rotatebox{40}{\makecell{Flowers102}} &
\rotatebox{40}{\makecell{DTD}} &
\rotatebox{40}{\makecell{EuroSAT}} &
\rotatebox{40}{\makecell{FGVC\\Aircraft}} &
\rotatebox{40}{\makecell{Caltech-101}} &
\rotatebox{40}{\makecell{Caltech-256}} &
\rotatebox{40}{\makecell{StanfordCars}} &
\rotatebox{40}{\makecell{PCAM}} &
\rotatebox{40}{\makecell{Average}} \\
\midrule
CLIP & 0.88 & 2.42 & 0.26 & 26.11 & 1.00 & 6.60 & 3.84 & 1.19 & 2.02 & 0.05 & 0.00 & 19.88 & 12.60 & 0.20 & 0.11 & 5.14 \\
FT-Clean & 13.55 & 19.92 & 4.94 & 40.00 & 0.82 & 2.64 & 2.40 & 0.68 & 2.66 & 0.05 & 0.03 & 14.95 & 9.69 & 0.09 & 1.32 & 7.58 \\
FT-Adv. & \underline{51.59} & 38.58 & 21.28 & 69.55 & 17.60 & 12.55 & 34.97 & 19.92 & \underline{15.90} & 11.95 & \underline{1.83} & 50.73 & \underline{48.48} & 8.42 & \textbf{48.88} & 30.15 \\
TeCoA & 37.57 & 30.30 & 17.53 & 69.17 & 19.70 & \underline{14.76} & 36.44 & 22.46 & 17.45 & \underline{12.14} & 1.62 & 55.86 & 41.89 & 8.79 & 47.39 & 28.87 \\
FARE & 23.88 & 21.25 & 10.72 & 59.59 & 8.30 & 10.97 & 24.56 & 15.48 & 10.96 & 0.14 & 0.84 & 45.96 & 34.35 & 4.38 & 10.17 & 18.77 \\
PMG-AFT & 47.11 & \underline{46.01} & \underline{25.83} & \underline{73.92} & \textbf{22.21} & 19.58 & \underline{41.62} & \underline{23.45} & 15.05 & \textbf{12.54} & 1.98 & \underline{62.42} & 45.99 & \underline{11.72} & \underline{48.64} & \underline{33.20} \\
\midrule
CAW & \textbf{52.16} & \textbf{48.21} & \textbf{27.99} & \textbf{74.83} & \underline{21.33} & \textbf{22.72} & \textbf{43.41} & \textbf{24.06} & \textbf{18.24} & 11.93 & \textbf{3.51} & \textbf{63.99} & \textbf{48.68} & \textbf{14.68} & 47.92 & \textbf{34.91} \\
\bottomrule
\end{tabular}
\end{adjustbox}
\end{table*}
% =========================
% Table 3
% =========================
\begin{table*}[!htbp]
\centering
\caption{Zero-shot clean accuracy under PGD-100 with $\epsilon = 1/255$ on 15 datasets, fine-tuned on TinyImageNet using PGD-2.}
\label{tab:clean}

\scriptsize
\renewcommand{\arraystretch}{1.1}
\setlength{\tabcolsep}{3pt}

\begin{adjustbox}{max width=\textwidth}
\begin{tabular}{l*{16}{c}}
\toprule
\textbf{Methods} &
\rotatebox{40}{\makecell{Tiny-ImageNet}} &
\rotatebox{40}{\makecell{CIFAR-10}} &
\rotatebox{40}{\makecell{CIFAR-100}} &
\rotatebox{40}{\makecell{STL-10}} &
\rotatebox{40}{\makecell{SUN397}} &
\rotatebox{40}{\makecell{Food101}} &
\rotatebox{40}{\makecell{OxfordPets}} &
\rotatebox{40}{\makecell{Flowers102}} &
\rotatebox{40}{\makecell{DTD}} &
\rotatebox{40}{\makecell{EuroSAT}} &
\rotatebox{40}{\makecell{FGVC\\Aircraft}} &
\rotatebox{40}{\makecell{Caltech-101}} &
\rotatebox{40}{\makecell{Caltech-256}} &
\rotatebox{40}{\makecell{StanfordCars}} &
\rotatebox{40}{\makecell{PCAM}} &
\rotatebox{40}{\makecell{Average}} \\
\midrule
CLIP & 57.26 & \textbf{88.06} & 60.45 & \textbf{97.04} & \textbf{57.26} & \textbf{83.89} & \textbf{87.41} & \textbf{65.47} & \textbf{40.69} & \textbf{42.59} & \textbf{20.25} & \textbf{85.34} & \textbf{81.73} & \textbf{52.02} & \textbf{52.09} & \textbf{64.77} \\
FT-Clean & \textbf{79.04} & 84.55 & 54.25 & 93.78 & 46.80 & 80.98 & 46.33 & 30.32 & 24.39 & 9.30 & 9.30 & 78.69 & 70.81 & 31.15 & 47.89 & 52.51 \\
FT-Adv & 73.83 & 68.96 & 39.69 & 86.89 & 33.37 & 27.74 & 60.10 & 33.45 & 13.26 & 16.49 & 4.86 & 67.41 & 57.72 & 18.11 & 49.91 & 43.45 \\
TeCoA & 63.97 & 66.14 & 36.74 & 87.24 & 40.54 & 35.11 & 66.15 & 33.25 & 13.75 & 17.13 & 6.75 & 64.63 & 56.20 & 25.65 & 49.01 & 44.15 \\
FARE & 77.54 & 87.58 & \textbf{62.80} & 94.33 & 49.91 & 70.02 & 81.47 & 57.10 & 36.33 & 22.69 & 14.19 & 84.04 & 77.50 & 44.35 & 46.07 & 60.39 \\
PMG-AFT & 67.11 & 74.62 & 44.68 & 88.85 & 37.42 & 37.47 & 66.34 & 35.66 & 21.17 & 17.76 & 4.71 & 76.70 & 61.96 & 25.21 & 49.60 & 47.28 \\
\midrule
CAW & 75.64 & 82.96 & 55.49 & 91.36 & 41.96 & 50.87 & 71.02 & 42.15 & 28.56 & 23.42 & 9.42 & 80.66 & 67.94 & 34.88 & 49.98 & 53.75 \\
\bottomrule
\end{tabular}
\end{adjustbox}
\end{table*}
% =========================
% Table 4
% =========================
\begin{table*}[!htbp]
\centering
\caption{Zero-shot robust accuracy under CW attack on 15 datasets. All methods are fine-tuned on TinyImageNet using PGD-2.}
\label{tab:cw}

\scriptsize
\renewcommand{\arraystretch}{1.1}
\setlength{\tabcolsep}{3pt}

\begin{adjustbox}{max width=\textwidth}
\begin{tabular}{l*{16}{c}}
\toprule
\textbf{Methods} &
\rotatebox{40}{\makecell{Tiny-ImageNet}} &
\rotatebox{40}{\makecell{CIFAR-10}} &
\rotatebox{40}{\makecell{CIFAR-100}} &
\rotatebox{40}{\makecell{STL-10}} &
\rotatebox{40}{\makecell{SUN397}} &
\rotatebox{40}{\makecell{Food101}} &
\rotatebox{40}{\makecell{OxfordPets}} &
\rotatebox{40}{\makecell{Flowers102}} &
\rotatebox{40}{\makecell{DTD}} &
\rotatebox{40}{\makecell{EuroSAT}} &
\rotatebox{40}{\makecell{FGVC\\Aircraft}} &
\rotatebox{40}{\makecell{Caltech-101}} &
\rotatebox{40}{\makecell{Caltech-256}} &
\rotatebox{40}{\makecell{StanfordCars}} &
\rotatebox{40}{\makecell{PCAM}} &
\rotatebox{40}{\makecell{Average}} \\
\midrule
CLIP & 0.21 & 0.36 & 0.10 & 10.59 & 1.16 & 0.82 & 1.23 & 1.09 & 2.18 & 0.01 & 0.00 & 13.50 & 7.36 & 2.36 & 0.07 & 2.45 \\
PMG-AFT & 44.59 & 44.86 & 24.15 & 74.11 & 19.99 & 17.33 & 39.88 & 20.95 & 13.51 & \textbf{12.09} & 1.47 & 60.99 & 44.46 & 10.57 & \textbf{48.59} & 32.36 \\
\midrule
CAW & \textbf{51.70} & \textbf{47.68} & \textbf{26.80} & \textbf{74.62} & \textbf{20.46} & \textbf{21.52} & \textbf{43.79} & \textbf{22.29} & \textbf{16.22} & 11.60 & \textbf{3.51} & \textbf{63.48} & \textbf{47.91} & \textbf{14.09} & 47.71 & \textbf{34.87} \\
\bottomrule
\end{tabular}
\end{adjustbox}
\end{table*}
\paragraph{Confidence Analysis}  
To understand how our proposed method (CAW) influences model predictions, we analyze the relationship between clean confidence and robust confidence for two datasets, TinyImageNet and CIFAR-10, as shown in Figure~\ref{fig:confidence_pgd_aa}. Here, we define confidence as the probability assigned to the true class by the model, i.e., the predicted probability corresponding to the ground-truth label. Specifically, clean confidence refers to this probability when computed on the clean (unperturbed) image, while robust confidence refers to the probability assigned to the true class for the corresponding adversarially perturbed image. Each point in the figure represents the confidence of a sample before and after fine-tuning, under two adversarial attacks: PGD-100 (right) and AutoAttack (left). In these plots, green points represent samples correctly classified in both clean and adversarial forms, orange points correspond to samples correctly classified only in the clean case, blue points (rare) denote samples correctly classified only in the adversarial version, and red points represent samples misclassified in both cases.

Before fine-tuning, most samples (orange and red) lie near the x-axis, indicating low robust confidence even when clean confidence is high. These ``Clean Only'' points demonstrate that the model fails to transfer confidence from clean samples to their adversarial counterparts. Only a few samples (green) are correctly classified in both cases, showing that small perturbations can easily change the model’s predictions. This pattern is consistent across both datasets.

After fine-tuning with CAW, the distribution shifts toward the diagonal. The increase in green points illustrates the effectiveness of our method in improving robustness. The positive correlation between clean and adversarial confidence indicates enhanced stability and resilience to perturbations. Additionally, the number of ``Clean Only'' points decreases, reflecting increased robustness. Notably, the red points remain near the origin, showing that CAW avoids overconfident predictions on misclassified samplesespecially under AutoAttack, and instead enhances robustness selectively where it is most beneficial.

This analysis under both attack types shows that CAW effectively stabilizes the model's predictions on low-confidence samples that are more vulnerable to adversarial perturbations. In other words, the model learns a more global and robust solution. Although CAW is fine-tuned using PGD-2 adversarial examples, it generalizes well to both PGD-100 and AutoAttack, demonstrating that CAW improves robustness without overfitting to the specific attack used during training.
\begin{figure*}[!t]
  \centering
  \includegraphics[width=0.98\textwidth]{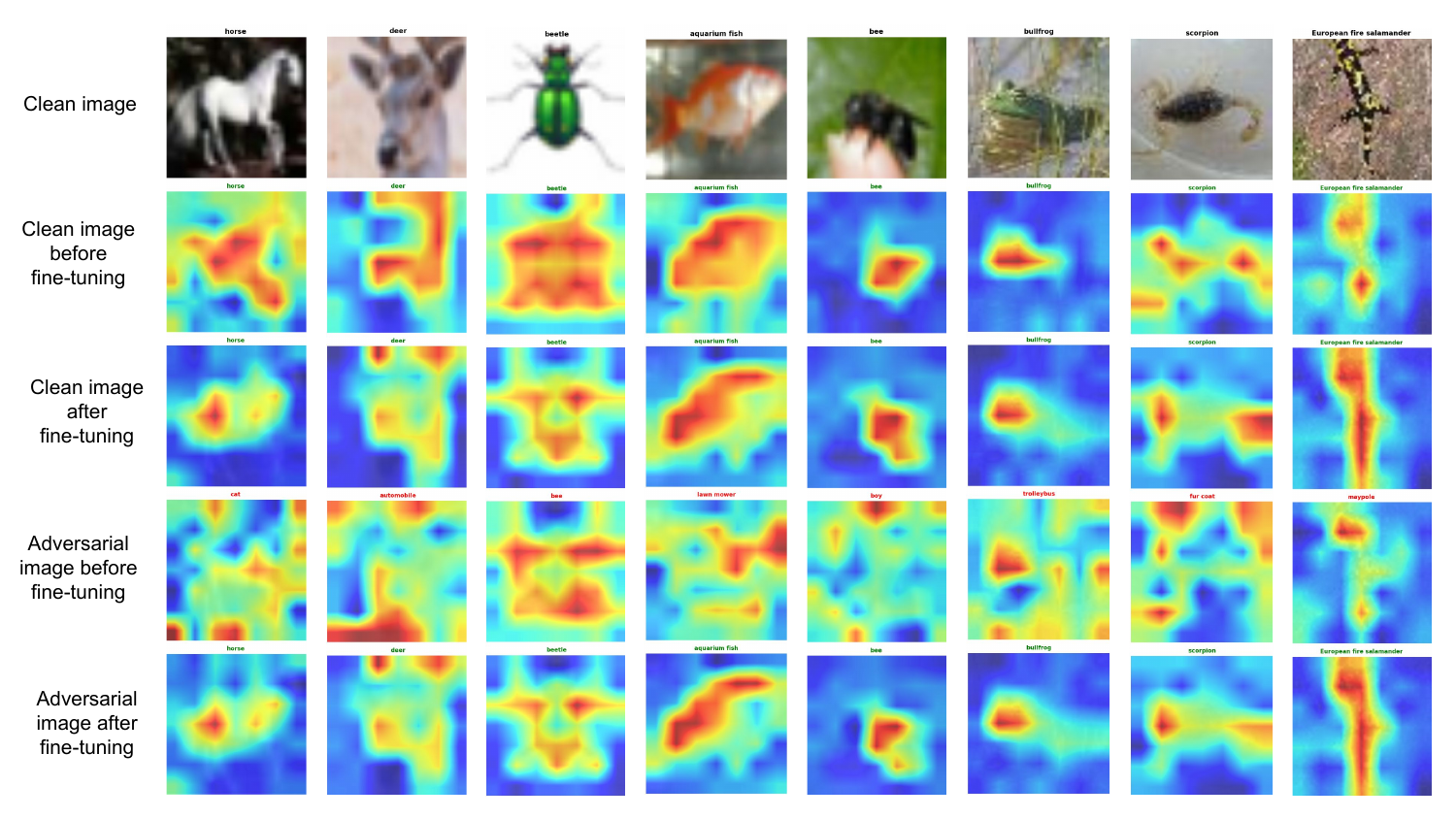}% no 1\textwidth
  \caption{Visualization of attention maps on clean and adversarial samples across CIFAR-10 (first two columns), CIFAR-100 (middle three), and TinyImageNet (last three), shown before and after fine-tuning with CAW. Labels are displayed at the top of each image and attention map: black indicates the ground-truth label, green indicates a correct prediction, and red indicates an incorrect one.}
  \label{fig:attention}
\end{figure*}
\paragraph{Attention Map Analysis}
Figure~\ref{fig:attention} illustrates the effect of CAW on attention map across clean and adversarial samples from multiple datasets. For each sample, we visualize attention maps on the clean image and its adversarial counterpart, both before and after fine-tuning with CAW. Each attention map is shown along with its predicted label at the top, where green indicates a correct prediction and red indicates an incorrect one.

Before fine-tuning, the attention maps for adversarial examples are often scattered or focused on irrelevant regions, such as the background or semantically incorrect parts of the image. This indicates that the model is easily distracted by small perturbations and struggles to attend to meaningful features. Even for clean images, the attention sometimes appears weak or inconsistent.

After fine-tuning with CAW, the attention maps become significantly more focused and semantically meaningful. For clean images, attention consistently centers on object-relevant regions, enhancing interpretability. For adversarial examples, the attention patterns shift to closely resemble those of their clean counterparts, suggesting that CAW helps the model maintain consistent internal representations even under attack. 

These visualizations support the conclusion that CAW not only improves robustness but also helps the model focus on semantically important features in both clean and adversarial examples.

\subsection{Ablation studies}
\paragraph{Effect of Attack Strength}  
Table~\ref{tab:attack_strength} presents the average robust accuracy under PGD-100 with perturbation bounds of $\epsilon = 1/255$, $2/255$, and $4/255$ across 15 datasets. Our method outperforms PMG-AFT on average and surpasses other baseline methods across various attack strengths.
% =========================
% Table 5
% =========================
\begin{table*}[!htbp]
\centering
\caption{Average zero-shot robust accuracy under PGD-100 with $\epsilon = 1/255$, $2/255$, and $4/255$, fine-tuned on TinyImageNet using PGD-2.}
\label{tab:attack_strength}

\scriptsize
\renewcommand{\arraystretch}{1.1}
\setlength{\tabcolsep}{3pt}

\begin{adjustbox}{max width=\textwidth}
\begin{tabular}{l*{16}{c}}
\toprule
\textbf{Methods} &
\rotatebox{40}{\makecell{Tiny-ImageNet}} &
\rotatebox{40}{\makecell{CIFAR-10}} &
\rotatebox{40}{\makecell{CIFAR-100}} &
\rotatebox{40}{\makecell{STL-10}} &
\rotatebox{40}{\makecell{SUN397}} &
\rotatebox{40}{\makecell{Food101}} &
\rotatebox{40}{\makecell{OxfordPets}} &
\rotatebox{40}{\makecell{Flowers102}} &
\rotatebox{40}{\makecell{DTD}} &
\rotatebox{40}{\makecell{EuroSAT}} &
\rotatebox{40}{\makecell{FGVC\\Aircraft}} &
\rotatebox{40}{\makecell{Caltech-101}} &
\rotatebox{40}{\makecell{Caltech-256}} &
\rotatebox{40}{\makecell{StanfordCars}} &
\rotatebox{40}{\makecell{PCAM}} &
\rotatebox{40}{\makecell{Average}} \\
\midrule
CLIP     & 0.64 & 2.15 & 0.12 & 20.35 & 0.52 & 5.94 & 2.97 & 0.72 & 0.71 & 0.03 & 0.00 & 14.28 & 9.18 & 0.11 & 0.04 & 3.65 \\
FT-Clean & 12.44 & 18.80 & 4.65 & 37.16 & 0.43 & 0.52 & 2.03 & 0.41 & 0.92 & 0.02 & 0.01 & 13.02 & 7.96 & 0.03 & 0.44 & 6.21 \\
FT-Adv   & \underline{29.33} & 18.10 & 11.06 & 45.13 & 8.58 & 5.65 & 16.45 & 10.15 & 9.72 & 9.82 & 0.83 & 33.43 & 24.14 & 3.80 & \underline{38.06} & 17.07 \\
TeCoA    & 18.17 & 12.78 & 8.12 & 39.87 & 8.53 & 6.12 & 11.04 & 10.07 & 10.07 & \underline{9.88} & 0.63 & 34.94 & 23.92 & 3.45 & 33.20 & 15.41 \\
FARE     & 12.41 & 9.09 & 4.23 & 33.72 & 2.98 & 4.75 & 9.67 & 5.52 & 4.26 & 0.25 & 0.28 & 23.97 & 16.95 & 1.48 & 3.43 & 8.54 \\
PMG-AFT  & 25.30 & 21.71 & 13.29 & 47.69 & \textbf{11.42} & 9.49 & \textbf{20.68} & \textbf{12.86} & 9.45 & \textbf{10.65} & \underline{0.90} & \textbf{41.86} & 28.92 & 3.72 & \textbf{37.88} & \underline{19.27} \\
\midrule
CAW      & \textbf{31.15} & \textbf{22.49} & \textbf{13.67} & \textbf{47.99} & \underline{9.87} & \textbf{9.88} & \underline{20.16} & \underline{12.14} & \textbf{10.90} & 7.05 & \textbf{1.43} & \underline{41.33} & \textbf{29.06} & \textbf{6.03} & 29.88 & \textbf{19.53} \\
\bottomrule
\end{tabular}
\end{adjustbox}
\end{table*}
% \begin{figure}[t]
%   \centering
%   \includegraphics[width=\columnwidth]{aa.png}
%   \caption{confidence of sample under autoattack attack.}
%   \label{fig:confidence_aa}
% \end{figure}
% \label{subsec:ablation}
\paragraph{Analyzing the Effect of Each Loss Component}  
As shown in Table~\ref{tab:ablation_cost}(a), the ${L}_{CE}$ row reports the average clean and robust accuracy across all 15 datasets under PGD-100 with $\epsilon = 1/255$. The ${L}_{CA}$ row presents the results after adding this component to the previous loss term. Finally, the ${L}_{Reg}$ row reflects the performance using the full loss function. These results demonstrate that our method improves both robustness and clean accuracy on average, compared to the standard CLIP loss.

\paragraph{Analysis of Computational Cost and Memory Usage}
As shown in Table~\ref{tab:ablation_cost}(b), our method uses less memory than both PMG-AFT and TGA-ZSR while achieving better accuracy under stronger attacks, as discussed in previous sections. It also maintains a training time comparable to the aforementioned approaches.
% =========================
% Combined Table: Ablation + Cost
% =========================
\begin{table}[!htbp]
\centering
\caption{Ablation study and computational cost analysis.}
\label{tab:ablation_cost}

\begin{minipage}[t]{0.46\textwidth}
\vspace{0pt}
\centering
\textbf{(a) Component Analysis}

\vspace{6pt}
\footnotesize
\setlength{\tabcolsep}{4pt}
\begin{tabular}{lccc}
\toprule
Loss Term & Robust & Clean & Average \\
\midrule
CLIP & 4.90 & 64.42 & 34.66 \\
\addlinespace
$\mathcal{L}_{CE}$ & 30.39 & 45.58 & 37.98 \\
$+\mathcal{L}_{CA}$ & 33.64 & 51.50 & 42.57 \\
$+\mathcal{L}_{Reg}$ & \textbf{34.92} & \textbf{53.65} & \textbf{44.28} \\
\bottomrule
\end{tabular}
\end{minipage}
\hfill
\begin{minipage}[t]{0.50\textwidth}
\vspace{0pt}
\centering
\textbf{(b) Memory and Training Cost}

\vspace{6pt}
\footnotesize
\setlength{\tabcolsep}{4pt}
\begin{tabular}{lcc}
\toprule
Methods & \makecell{Memory\\(MB)} & \makecell{Time\\(epoch/batch)} \\
\midrule
CLIP & 0 & 0s / 0s \\
TeCoA & 12{,}873 & 512s / 0.65s \\
CAW & 15{,}986 & 842s / 1.08s \\
PMG-AFT & 18{,}449 & 828s / 1.06s \\
TGA-ZSR & 21{,}227 & 885s / 1.13s \\
\bottomrule
\end{tabular}
\end{minipage}

\end{table}
\FloatBarrier
\section{Conclusion and Limitations}

In this work, we demonstrate that emphasizing vulnerable samples during training improves the zero-shot robustness of CLIP. To this end, we introduce a CAW method that encourages the model to focus on hard adversarial examples, enabling the learning of more robust and transferable features. Experimental results show that our method outperforms prior approaches in both clean and robust accuracy across diverse domains under strong attacks, while requiring less memory, which is important for large-scale models. For future work, we aim to design a loss function that combines the idea of weighting challenging samples with attention mechanisms, which are essential components of large-scale models, to achieve better robustness against various attacks. Additionally, improving model interpretability by analyzing the features that contribute to robustness on difficult examples may provide deeper insights into building more resilient vision-language models.
\paragraph{Limitations and Broder impact }
Our method focuses solely on the CLIP model and has not been tested on other vision-language models under adversarial attacks. In addition, it only addresses adversarial perturbations in the image encoder, whereas the text encoder is also a crucial component of VLMs and should be considered to improve overall robustness. Large-scale models like VLMs have demonstrated strong zero-shot capabilities, performing well across diverse tasks and unseen domains. However, their performance under adversarial perturbations remains limited, which is an important and active area of research. As these models are increasingly deployed in real-world applications, ensuring their robustness and privacy against adversarial attacks becomes critical. Our method aims to improve the zero-shot robustness of CLIP under such attacks, contributing to the development of safer and more reliable VLMs. Furthermore, our evaluation focuses on standard white-box attacks. Black-box attacks, adaptive attacks, and larger perturbation radii were not considered in this work and remain important directions for future investigation.

\section*{CRediT authorship contribution statement}

\textbf{Nikoo Naghavian:} Conceptualization, Methodology, Software, Investigation, Formal analysis, Validation, Visualization, Writing – original draft, Writing – review \& editing.

\textbf{Mostafa Tavassolipour:} Conceptualization, Methodology, Supervision, Validation, Writing – review \& editing.

\begin{comment}
For transparency, we require corresponding authors to provide co-author contributions to the manuscript using the relevant CRediT roles. The CRediT taxonomy includes 14 different roles describing each contributor’s specific contribution to the scholarly output. The roles are: Conceptualization; Data curation; Formal analysis; Funding acquisition; Investigation; Methodology; Project administration; Resources; Software; Supervision; Validation; Visualization; Roles/Writing - original draft; and Writing - review & editing. Note that not all roles may apply to every manuscript, and authors may have contributed through multiple roles.
\end{comment}

\section*{Declaration of Competing Interest}
The authors declare that they have no known competing financial interests or personal relationships that could have appeared to influence the work reported in this paper.

% \section*{Acknowledgements}
% Collate acknowledgements in a separate section at the end of the article before the references and do not, therefore, include them on the title page, as a footnote to the title or otherwise. List here those individuals who provided help during the research (e.g., providing language help, writing assistance or proof reading the article, etc.).

\section*{Data availability}
The datasets used in this study are publicly available from their respective official sources. No new dataset was created during this work.

\bibliography{sample}

\end{document}